# Constructing Epistemic AI Literacy: Detecting Epistemic Aims and Processes in Student-AI Co-Programming*

Mengqian Wu[1]

[1]McGill University, Quebec, Canada

## Abstract

Epistemic thinking plays a central role in students' learning processes when applying generative artificial intelligence (GenAI), particularly in programming contexts where learners must construct queries, evaluate and validate AI-generated outputs, and regulate problem-solving strategies. This study introduces observable the conceptual framework of Epistemic AI Literacy (EAIL), reframing AI literacy as a process-oriented epistemic phenomenon that emerges through dynamic human-AI interactions across different domains. Drawing on AIR (epistemic aims, ideals and reliable epistemic processes) framework, this study examines how epistemic aims and epistemic processes are enacted in GenAI-supported co-programming activities and explores scalable approaches for operationalizing these constructs in interaction data. Using a large dialogue dataset of human–AI co-programming, this study identifies observable dimensions of epistemic aims (i.e., mastery-oriented aims) and epistemic processes (i.e., outsourcing, explanation seeking, verification seeking, prompt monitoring, and epistemic justification). A subset of interactions is manually annotated to ground these constructs, which then inform scalable automatic labeling using complementary approaches interactively including few-shot prompting and regex-based scripts. The results reveal a prevalent lack of EAIL, with 78.8% of student-GenAI interactions relying on non-mastery-oriented aims and less reliable epistemic strategies like outsourcing and verification-seeking. Conversely, only 11.1% of interactions showed high epistemic engagement, where mastery-oriented aims were coupled with advanced epistemic strategies like epistemic justification in a more reliable epistemic process. These findings suggest that while GenAI facilitates task success, robust epistemic performance and genuine learning rarely emerges without deliberate instructional and design support.

## 1. Introduction

According to a 2022 UNESCO report [1], 'a technology-oriented approach has been typically taken towards AI skills training, and AI is usually only taught as part of the computing curriculum. Human and in-depth ethical questions are too often ignored' (p. 14). It highlights a central limitation of prevailing AI education approaches. When framing skills in AI literacy as technical competencies, it is common to overlook the human, epistemic, and ethical dimensions of how AI is actually used in practice. When instruction focuses mainly on how AI systems function, rather than how people interpret, trust, evaluate, and regulate AI-generated knowledge, learners are left unprepared to engage critically with AI in real decision-making contexts.

Fostering students' AI literacy is incrementally essential, specifically cultivating a specific set of skills and knowledge related to understanding and engaging with AI technologies [2]. AI literacy is broadly defined as learners' ability to understand, use, evaluate, and critically engage with AI systems across technical, ethical, and societal dimensions [3][4][5]. Research on AI literacy has expanded significantly in recent years, but ongoing debate continues concerning whether literacy should be treated as general awareness or as a measurable competency, with many frameworks operationalizing literacy in competency-like ways to make it actionable [3][6][7]. Accordingly, existing AI literacy research has developed a range of frameworks and assessment instruments, most commonly operationalized through self-report questionnaires and multiple-choice knowledge tests [8][9]. For example, the ABCE framework [4] serves as a prominent model for AI literacy, with its cognitive dimensions closely aligned with the hierarchical structure of Bloom's taxonomy [10]. Within this framework, cognitive learning is categorized across a spectrum of lower- to higher-order thinking skills, reflecting the progression of intellectual development [11]. Most existing scales have not systematically examined measurement error, floor effects, and ceiling effects [12], and have been critiqued for insufficient attention to problem-solving in AI-assisted contexts [13].

*Corresponding author.
mengqian.wu@mail.mcgill.ca (M. Wu)
0000-0002-7217-6158 (M. Wu)

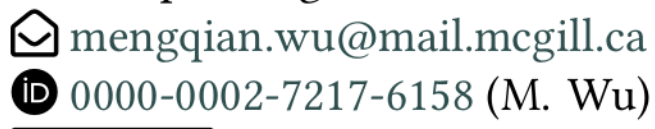

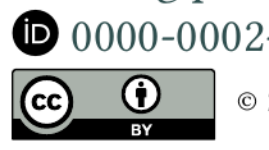

Although these approaches have contributed to a baseline understanding of learners' AI-related knowledge, skills and attitudes, they tend to conceptualize AI literacy as a stable, trait-like construct, largely independent of context and human-AI interaction. These measurement instruments are not necessarily capturing real literacy or competency, because they primarily capture lower-order cognitive outcomes (e.g., factual knowledge, recognition of AI concepts, or self-perceived skills) and underrepresent higher-order thinking skills such as evaluation, justification, verification, and regulation. In addition, trait-based measures often simplify the cognitive and epistemic demands of real human–AI interaction, abstracting away from the dynamic, iterative, and context-specific nature of how learners actually work with genAI tools in authentic tasks (e.g., programming, domain knowledge inquiry). As a result, such measures risk overestimating learners' AI literacy by capturing surface-level success rather than deeper understanding or transferable competence.

To address these gaps, this study introduces *Epistemic AI Literacy* (EAIL), defined as: the ability to understand, regulate, and critically engage with AI systems as cognitive and decision-making agents, specifically through how they pursue, evaluate, and regulate knowledge when interacting with AI systems. This includes the capacity to allocate, monitor and reclaim cognitive and knowledge authority when collaborating with AI, such as determining when to trust, outsource, verify, justify, and override AI-generated judgments. Operationalizing this concept enables the simultaneous construction of an entirely new theoretical and measurement framework. On this basis, the study first establishes an EAIL framework to evaluate students' real-time epistemic and cognitive processes during human-AI interaction. Subsequently, these detection methods are scaled by examining the theoretical grounding and analytic robustness of multiple automatic labeling approaches to augment human annotation. This study investigates how students engage epistemically with generative AI during programming tasks: (RQ1) examines the epistemic aims and epistemic processes that students enact when interacting with AI in authentic co-programming contexts; (RQ2) evaluates how effectively different automatic labeling approaches (with/without regex-based rules in few-shots learning) detect these epistemic aims and processes; (RQ3) and shows how the resulting labels reflecting their distribution and clustering patterns as different profiles of epistemic performance within student–AI dialogue. Together, these research questions address both the substantive nature of students' epistemic engagement with AI and the methodological feasibility of scaling its analysis in large interaction datasets. This research offers new perspectives on the construction of conceptual frameworks and measurement systems for AI literacy. It further demonstrates that EAIL indicators can serve as critical signals for identifying deficiencies in epistemic engagement, as a core condition for achieving genuine learning and avoiding superficial pseudo-success.

### 1.1. Epistemic AI Literacy Framework

To justify integrating epistemic thinking frameworks with AI literacy, the core argument is that when AI literacy is treated as a set of skills or competencies, it is best observed in real human AI interaction rather than measured only through stable, trait-like scales. Such scales often fail to capture higher order and complex thinking processes. In the era of GenAI, AI literacy is most clearly enacted during direct interaction with AI systems, especially when learners engage in tasks that require higher order thinking. In these situations, learners must formulate prompts, interpret AI responses, evaluate their relevance and correctness, verify claims, and regulate subsequent actions. Because students primarily interact with GenAI for knowledge seeking and knowledge construction, these activities are fundamentally epistemic in nature. They involve judgments about knowledge quality, justification, and reliability. As a result, understanding AI literacy requires attention not only to what learners know, but also to how they pursue, evaluate, and regulate knowledge when interacting with AI.

To achieve it, this study draws on theories of epistemic thinking. Epistemic thinking refers broadly to cognitive processes related to knowledge and knowing [14]. A key component of epistemic thinking is the pursuit of epistemic aims through reliable or unreliable processes [15]. Epistemic aims are defined as goals that focus on acquiring true and justified beliefs that are supported by sufficient reasons [16]. To make these processes measurable in learning contexts, it is more feasible to focus on small units of knowledge acquisition and problem solving. In human-GenAI co-programming activities, the small unit of analysis is the prompt-response pair. The belief formation and justification take the form of knowledge seeking and solution seeking through interactive inquiry guided by learner intentions. Although epistemic aims are not strictly categorized, insights from goal theory and achievement goal frameworks allow them to be operationalized using process based approaches that capture learners' epistemic activity as it unfolds during authentic AI supported tasks. While mastery-oriented goals prioritize personal growth and the acquisition of new skills, performance goals focus on validating one's competence through social comparison and outperforming peers [17]. This study classify student prompts that function as inquiry-based activities into mastery-oriented intent and non-mastery-oriented intent. This nuance is critical, because when working on programming tasks with GenAI students often focus on immediate task completion rather than long-term growth. It is different from a "performance" orientation, as they lack the social comparison typical of that goal. Labeling such task-focused behaviors strictly as 'performance-oriented' would be inaccurate.

Furthermore, epistemic processes refer to the cognitive and epistemic operations that facilitate knowledge construction, extending to mechanisms at strategy-level about how learners achieve epistemic aims based on their epistemic ideals [16]. For example, these processes can be observed through epistemic strategies such as validation and justification. Validation involves evaluating new information by comparing it with prior knowledge or checking its consistency with previously processed information [18]. Justification involves generating and evaluating reasons and evidence to support claims or solutions, including assessing the quality of others' reasoning [19]. In authentic learning contexts, epistemic processes vary in their reliability as mechanisms of knowledge construction, and analyzing both reliable and unreliable processes offers critical insight into students' epistemic engagement and the quality of their learning. Building on the established link between mastery-oriented goals and metacognition [20], it is important to investigate how epistemic aims are associated with the reliability, effectiveness, and variety of epistemic strategies learners employ. In the current study, a dual path inductive and deductive coding approach was used to identify five epistemic strategies that frequently appeared in student-GenAI interaction, including outsourcing, explanation seeking, verification seeking, prompt monitoring, and epistemic justification. Although the AIR components are interdependent, researchers are not required to examine all components simultaneously, and focusing on selected components is appropriate when aiming for in depth analysis [21]. Given the nature of the dataset, this study only focus on detecting and describing epistemic aims and epistemic processes. This decision allows for a focused and detailed analysis of learners' epistemic performance in AI-supported programming, preserving contextual sensitivity while contributing to generalizable theoretical insights.

Existing AI literacy frameworks emphasize the importance of higher order thinking such as analyzing, evaluating, and creating with AI systems [4]. These forms of thinking are central to meaningful epistemic engagement with GenAI. EAIL integrates a structure of epistemic thinking framework into a process-based evaluation of AI literacy that centers on epistemic engagement during human-GenAI interaction. In summary, EAIL addresses the growing need to move beyond trait-like measures toward process-based approaches that capture learners' nuanced cognitive and epistemic activity as it unfolds in authentic AI-supported tasks. Such approaches enable fine-grained analysis of moment-by-moment reasoning and decision making, providing a more valid representation of AI literacy in practice.

# 2. Data and Methods

## 2.1. Dataset Description

This study utilizes the StudyChat dataset [22], a publicly available corpus of naturalistic student–LLM interactions collected in an undergraduate AI course at a large U.S. research university. The dataset captures real-world usage of a custom GPT-4o powered virtual assistant designed to mirror ChatGPT's functionality while students worked on seven programming assignments over a full semester. The corpus consists of 1,197 unique conversation sessions containing approximately 7,856 student utterances. The interaction data are linked to programming assignment submissions, enabling fine-grained analyses of how specific inquiry behaviors relate to genuine learning versus surface-level success. For the present study, 200 complete co-programming chat sessions (approximately 2000 prompts) extracted through unique chat IDs were randomly sampled, each consisting of multiple pairs of student-LLM prompt-response turn pairs. From this subset, 499 turns were randomly selected for human annotation, while an additional 1,748 turns were labeled using GPT-4o to support assisted scaling, balancing pedagogical relevance with broader analytical coverage.

## 2.2. Measures and Evaluation

To evaluate the alignment between student inquiry and the proposed constructs, the analysis was performed in three distinct phases. The workflow is visualized in Figure 1. First, the AIR (aims, ideals, and reliable Processes) theoretical framework of epistemic thinking was utilized to operationalize epistemic aims and strategies within the specific context of human-AI co-programming. This framework provided the conceptual basis for distinguishing between task-oriented behaviors and deeper epistemic inquiries.

Second, a "gold standard" ground truth dataset was established through manual expert annotation. From a total corpus of approximately 2,000 conversation turns (prompt-response pairs), a representative sample of 499 lines was human-labeled. During this manual labeling process, experts iteratively developed and refined a suite of regex-based principles, including standardized linguistic patterns and heuristics that reliably indicated specific intents and cognitive processes of the students (see Table 1). For example, during the human annotation phase, a outsourcing strategy was identified, reflecting students' tendency to seek and trust GenAI outputs without engaging in questioning, interpretation, or justification. This strategy was operationalized as instances in which students submitted assignment questions, code snippets, or error messages to the AI without providing contextual explanation, interpretation, or a clearly articulated problem in the programming learning context. Prompts in this category typically present tasks, code, or errors with minimal reflective framing, signaling task outsourcing rather than active problem formulation. The central criterion for classification is whether the student demonstrates independent reasoning or problem analysis prior to engaging the AI. Based on these annotations, high-probability linguistic indicators were subsequently examined, and frequently occurring surface-level signals were distilled into regex-based rules capturing direct task delegation without problem framing, such as the presence of terms (e.g., "complete this," "solve the following," "do this problem," "help with," "fix," "make").

Finally, an automated labeling pipeline was implemented using GPT-4o through a few-shot learning. To validate the utility of the expert-derived heuristics, two experimental conditions were tested including 1) one group where the few-shot prompts included the regex-based principles as explicit guidance and 2) a control group where these regex-based principles were omitted. The performance of both automated approaches was evaluated against the 500-line human-labeled ground truth. Accuracy scores were calculated to measure the degree of agreement between LLM-generated labels and human expert annotations, effectively testing the reliability of LLM-supported classification in the programming context.

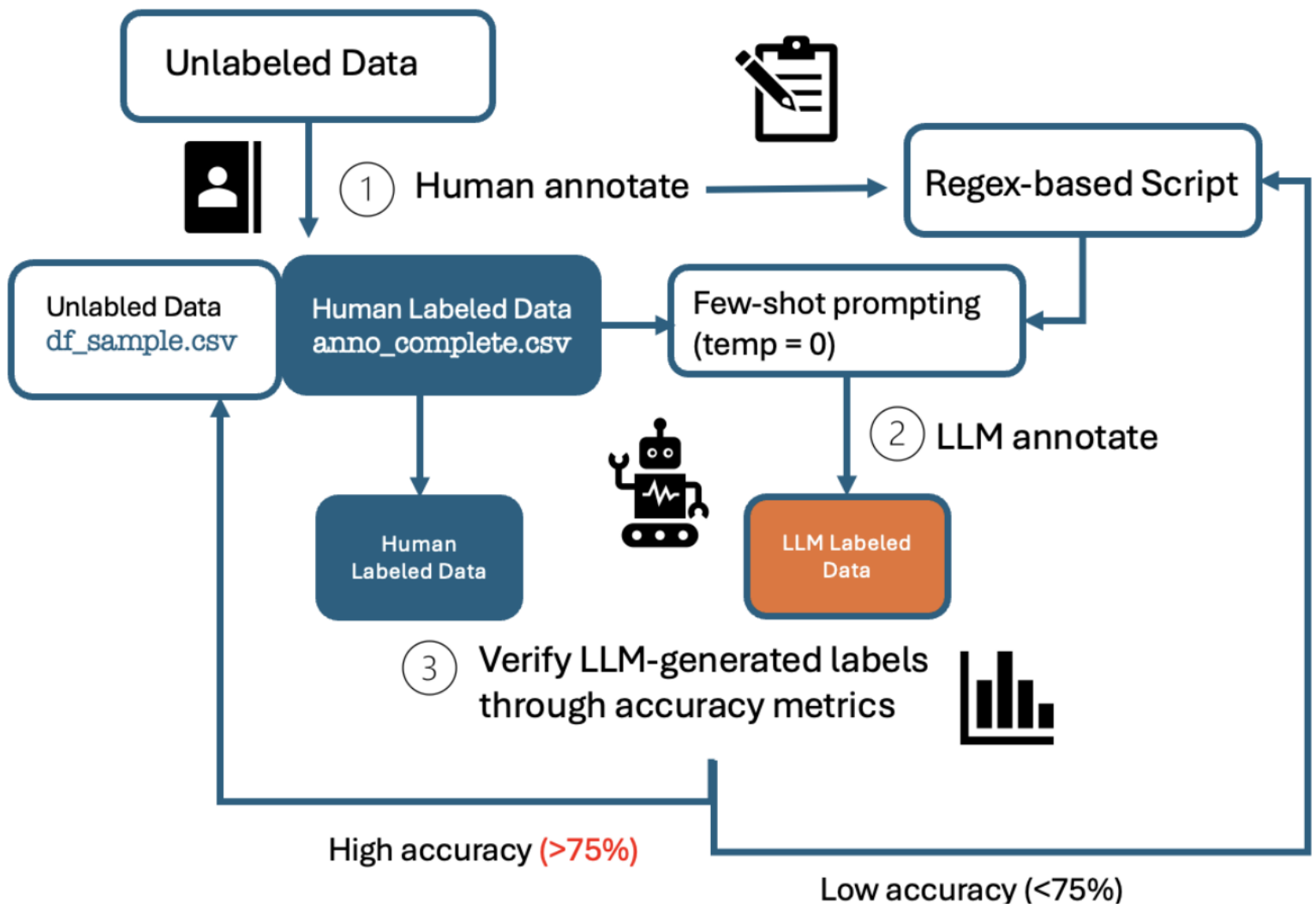

**Figure 1:** Automatic LLM labeling workflow with few-shot prompting and optional regex-based processing.

## 2.3. Regex-based Rules and LLM Few-Shot Prompt Pipeline

Regular regex-derived metrics are considered probabilistic lexical cues and predictors, potentially linked to underlying cognitive constructs rather than deterministic rules [23]. The use of lexical indicators to detect complex cognitive processes is often limited by a semantic gap, where context-sensitivity and diverse linguistic forms weaken the causal link between surface-level word usage and underlying intent [24][25]. Now researchers are increasingly leveraging LLMs for more nuanced and scalable automatic coding and qualitative codebook development [26]. One promising approach to automated labeling involves integrating code definitions, or label descriptions, with the linguistic knowledge acquired during the human annotation phase to serve as soft decision principles within few-shot prompting mechanisms. This allows models to leverage lexical signals while conditionally reweighting them based on conversational context and pragmatic intent. The design aims to preserve the informational value of lexical cues while reducing their fragility and improving the construct validity of discourse-level epistemic annotations. In this study, human–AI dialogue data were analyzed to identify seven essential analytic codes (see Table 1), including inquiry relevance, epistemic aims, and five epistemic strategies enacted during the inquiry process. Inquiry relevance captures whether a learner's prompt to the LLM is meaningfully related to the programming task or the core concepts taught in the computer science course. Prompts that are unrelated to the learning context (e.g., entertainment-oriented chat) are excluded, as epistemic engagement cannot be meaningfully examined in such cases. Epistemic aims are categorized as mastery-oriented or non-mastery-oriented. A mastery-oriented epistemic aim is inferred from prompts that signal an intention to understand underlying mechanisms, strategies, or reasoning processes, rather than merely requesting working code or surface-level solutions.

In addition, five epistemic strategies were identified from the data: (1) outsourcing, (2) explanation-seeking, (3) verification-seeking, (4) prompt monitoring, and (5) epistemic judgment. Outsourcing occurs when learners request direct answers from the LLM without engaging in problem framing or interpretation, often by pasting code, errors, or questions with minimal contextualization. This strategy reflects uncritical acceptance of LLM output and reliance on the model's authority, with little evidence of reflective or evaluative thinking. Explanation-seeking and verification-seeking strategies involve requests for conceptual or procedural understanding and for confirmation or debugging of one's own attempts (e.g., code, reasoning, or hypotheses), respectively. While these strategies are important for learning and reflect partially reliable epistemic processes, they do not necessarily indicate the learner's evaluative judgment toward the knowledge being generated. Prompt monitoring reflects learners' active regulation of the human–AI interaction by shaping how information is generated and delivered. This includes specifying stylistic or formatting constraints (e.g., concise, clear, step-by-step) and exercising temporal or pacing control (e.g., "wait," "let me think first"). Finally, epistemic judgment represents the most epistemically robust strategy, characterized by higher-order inquiry such as questioning assumptions, comparing alternative approaches, and probing trade-offs or conditional strategies. These prompts demonstrate reflective reasoning, epistemic curiosity, and evaluative engagement, for example when learners ask why one approach is preferable to another or request justification for a recommended solution.

**Table 1**
Construct mapping for epistemic AI literacy indicators used in our procedural labeling task.

| Construct | Definition | Ex. Indicator | Ex. Quotes |
|---|---|---|---|
| Relevance | Whether the prompt is related to learning CS knowledge or programming; if not relevant, all remaining indicators are set to 0. | Detecting disciplinary keywords in computer science or programming | "Can you wish me a happy birthday" |
| Mastery-oriented aims | Prompts signaling intent for understanding/strategy/reasoning (not just seeking answers for programming tasks), such as requiring explanations of causes, mechanisms, principles. | Using contrast words (e.g., 'but', 'however', 'differ', 'difference', 'better', 'easier', 'best', 'instead') to evaluate solutions or seek an alternative solution. | "Shouldn't it be dataset instead of dataframe? whats the difference between the two??" |
| Outsourcing | Delegating cognitive work without problem framing: pasting code/questions/errors with minimal interpretation or specific inquiry. | Direct task delegation without problem framing: (e.g., "complete this," "solve the following," "do this problem", "help with", "fix", "make") | "this is the question i'm supposed to answer." |
| Explanation seeking | Seeking conceptual or procedural understanding without primarily validating an existing solution. | Conceptual explanation requests using definition-seeking phrases (e.g., what is, what does it mean, could you explain) | "after training a logistic regression model how to predict the probability for each possible class" |
| Verification seeking | Requesting confirmation or debugging of the student's own attempt (e.g., code, reasoning, hypothesis). | Code verification requests including verification-oriented verbs or questions (e.g., check, verify, correct, confirm, diagnose, does this work) | "So with that information is my code sound without the change that you suggested to the base case?" |
| Prompt monitoring | Regulating the AI's behavior/output (style, scope, constraints) based on prior outputs or task needs. | Style and formatting constraints (e.g., concise, simple, clear, rearrange, stronger, write in 1–2 sentences, 3–4 lines, continuous paragraph); Interaction and process control using temporal or pacing directives (e.g., wait, slow down, take a second to think) | "Please start from the original code I provided for autocomplete.py and write all of the functions. Considering all of the code files I have provided please give me the correct functions. Take a second to think before doing this please" |
| Epistemic justification | Engaging in higher-order inquiry by questioning assumptions, comparing alternatives, or asking for reasons, trade-offs, or conditional strategies. These prompts reflect epistemic curiosity, agency, and reflective thinking, often extending beyond immediate task completion toward deeper understanding or generalization. | Reasoning and rationale seeking language: e.g., (why, intuition, make sense, justify, prove, show evidence, clarify, identify, essential/important) | "Why do we use `y_test` in `r2_using_score_method = model.score(X_test_poly, y_test)`?" |

For LLM-facilitated labeling, a small number of carefully curated examples, otherwise known as demonstrations, was adopted for few-shot prompting, with seven student prompts selected as a balance between representational coverage and cognitive economy. Prior work on in-context learning suggests that model performance is driven more by the quality and diversity of examples than by sheer quantity, with diminishing returns and potential over-conditioning observed as the number of demonstrations increases. Selecting seven examples allows the prompt to capture prototypical cases, boundary conditions, and negative contrasts within the label space, while preserving sufficient context length for the target interaction and reducing risks of instruction dilution or recency bias. This design aligns with empirical findings that a small number of demonstrations (e.g., on the order of a handful) is often sufficient to induce stable task-relevant reasoning patterns in LLMs[27]. In this study, we adopt seven representative examples for few-shot learning demonstrations, including most common patterns of target labels and special edge cases.

## 3. Results

**Regex-based rules empowers automatic labeling through LLM few-shot learning.** As shown in Table 2, few-shot LLM labeling is evaluated against the human-annotated gold set by matching records on `conversation_id` (499 overlapped conversations). Without regex-based rules, the LLM achieved an overall accuracy of 0.818 across seven binary dimensions. Adding lightweight regex-based rules improved overall accuracy to 0.852 (+0.035), yielding consistent gains on most dimensions, particularly *Mastery-oriented_aims* (0.709→0.835, +0.126), *Explanation-seeking* (0.788→0.839, +0.052), and *Outsourcing_process* (0.747→0.791, +0.044). Performance on *Inquiry_relevance* is stable (0.880→0.882), while *Verification-seeking* showed a small decrease (0.834→0.829, $-0.004$). Overall, these results suggest that simple rule-based constraints can meaningfully reduce common labeling errors in ambiguous categories while largely preserving performance on easier dimensions.

**Table 2**
Accuracy comparison of few-shot LLM labeling with vs. without regex-based rule processing.

| Dimension | No regex | With regex | $\Delta$ |
|---|---|---|---|
| Inquiry Relevance | 0.880 | 0.882 | $+0.002$ |
| Mastery-oriented Aims | 0.709 | 0.835 | $+0.126$ |
| Outsourcing | 0.747 | 0.791 | $+0.044$ |
| Explanation Seeking | 0.788 | 0.839 | $+0.052$ |
| Verification Seeking | 0.834 | 0.829 | $-0.004$ |
| Prompt Monitoring | 0.888 | 0.900 | $+0.012$ |
| Epistemic Justification | 0.878 | 0.890 | $+0.012$ |
| Overall Accuracy | 0.818 | 0.852 | $+0.035$ |

**Prevalence and distribution of seven binary labels.** Figure 2 provides a descriptive overview of the prevalence of seven binary labels including the inquiry relevance, epistemic aims (either mastery-oriented or non-mastery-oriented), and five epistemic strategies indicated previously. *Inquiry relevance* was present in the large majority of records (87.9%), indicating that most interactions were aligned with the inquiry goal in this computer science course during the semester. *Mastery-oriented aims* (21.2%) were present in a small proportion of records, suggesting that most students' prompts do not include explicit learning oriented goals or a focus on developing understanding. Epistemic processes strategy-related features occurred less consistently that *Outsourcing* and *Explanation Seeking* were each observed in approximately 35.0% of records, suggesting these were the most common strategy during epistemic processes. In contrast, *Verification Seeking* (20.5%) appeared in roughly one-fifth of the dataset, while *Epistemic Justification* was less frequent (13.3%). *Prompt Monitoring* was rare (3.2%), implying that explicit monitoring of the system prompt occurred only in a small subset of interactions. Overall, the figure highlights an imbalance in feature prevalence, motivating subsequent analyses of co-occurrence and profile patterns.

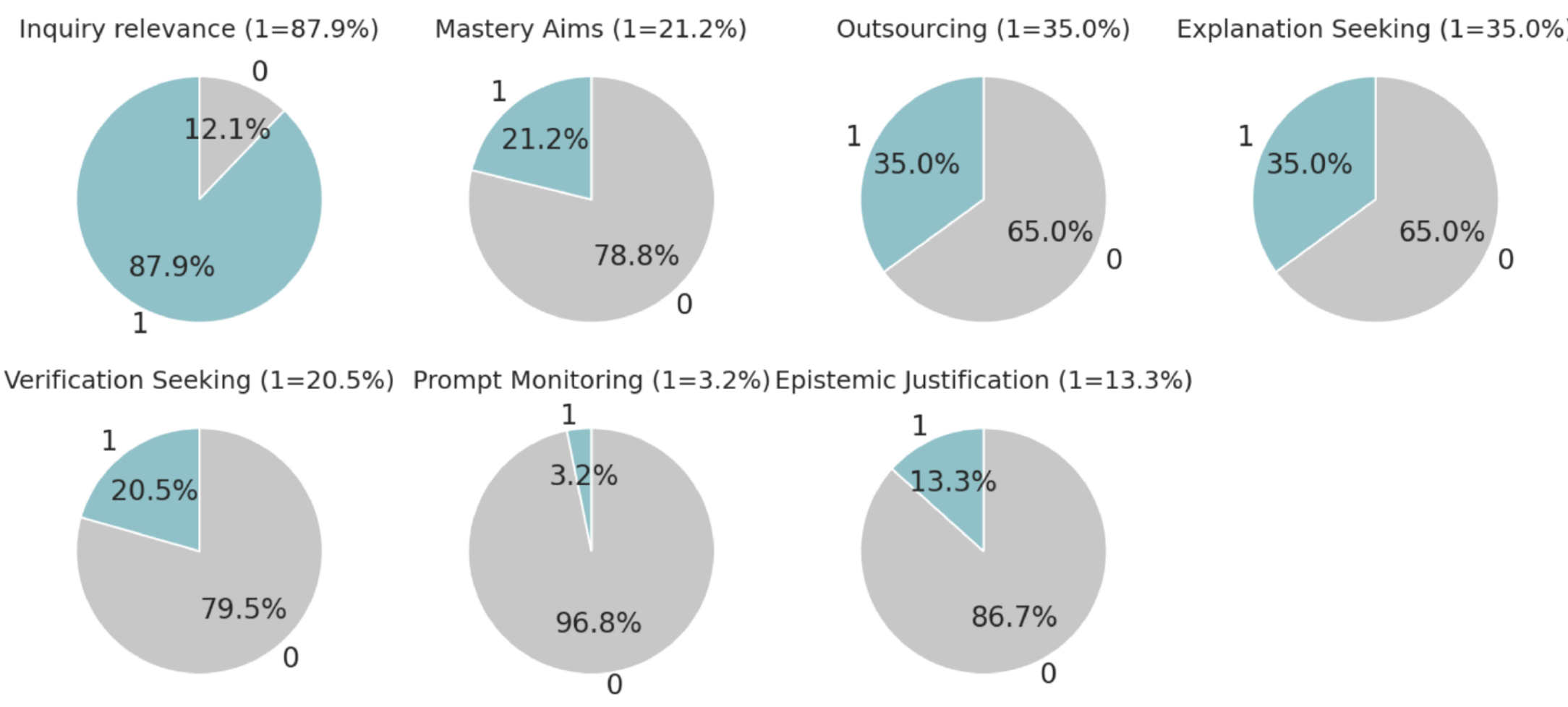

**Figure 2:** Distribution of seven binary labels and each pie chart shows the proportion of records coded as present (1) versus absent (0) for a given target label; For the mastery aims chart, it indicates mastery-oriented aims (1) versus non-mastery-oriented aims (0); percentages indicate the share of records with value 1.

As shown in Figure 3, the left panel (a) reports the $\phi$ correlation matrix among the seven binary labels, highlighting generally weak-to-moderate associations (e.g., a strong positive correlation between mastery-oriented aims and epistemic justification and a notable negative correlation between outsourcing and explanation seeking). The right panel (b) visualizes the ten most frequent profiles, where each row represents a common combination pattern across the same set of features.

**Top Ten Student EAIL Profiles After Clustering Epistemic Aims and Processes.** Figure 4 summarizes the ten most frequent binary-feature profiles observed in the student–GPT conversations. The distribution is highly concentrated in a small number of patterns, and the two most common profiles were *Inquiry relevance + Outsourcing* ($n = 403$, 25.0%) and *Inquiry relevance + Explanation Seeking* ($n = 348$, 21.6%), together accounting for 46.6% of all records. A significant share of records were coded as *Inquiry content not relevant* ($n = 212$, 13.2%), while the next most frequent relevant profiles combined inquiry relevance with *Verification Seeking* ($n = 157$, 9.7%) or with *Outsourcing + Verification Seeking* ($n = 150$, 9.3%). More elaborated profiles involving *Mastery Aims* (with *Explanation Seeking* and/or *Epistemic Justification*) were less common (7.3%–5.0%), and profiles featuring *Prompt Monitoring* appeared rarely (1.2%), indicating that relevance paired with outsourcing or explanation seeking dominated the observed interaction patterns. Taken together, outsourcing and explanation seeking reflect a pattern in which students rely on the LLM to directly generate solutions for programming tasks and frequently request explanations of concepts or procedures. Within the EAIL framework, these behaviors represent less reliable epistemic processes, as they do not consistently support optimal epistemic performance or genuine learning.

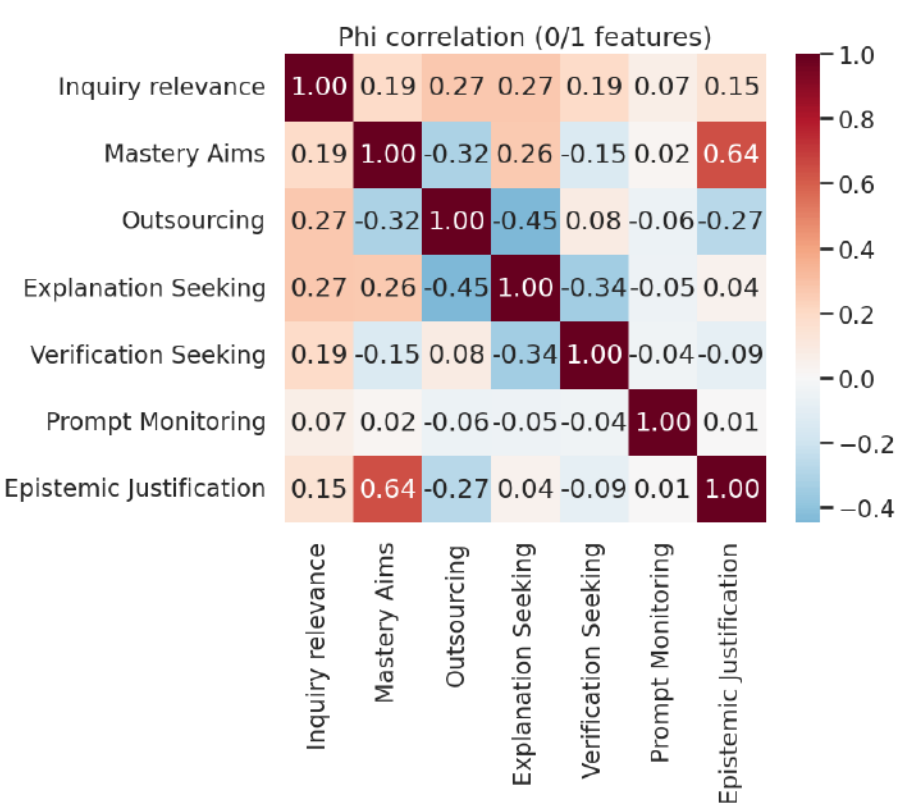

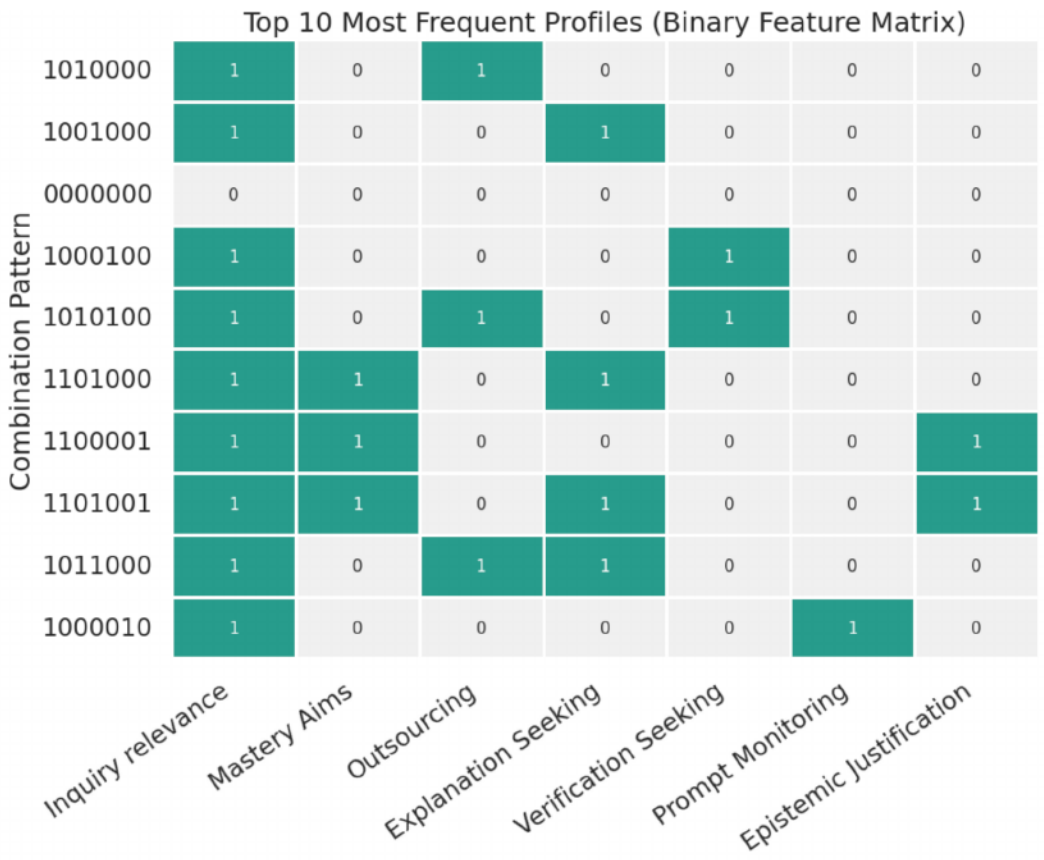

**Figure 3:** (a) Heatmap visualization of correlation among target variables. (b) Ten most frequent student profiles of EAIL performance combining relevance, epistemic aims (mastery-oriented or non-mastery-oriented), and epistemic strategies (outsourcing, explanation-seeking, verification-seeking, prompt monitoring, epistemic justification) applied in reliable/unreliable process.

Analysis of the five most frequent epistemic profiles suggests that these interaction patterns are unlikely to reflect epistemically productive engagement. In these profiles, learners rarely exhibit mastery-oriented epistemic aims and predominantly rely on less truth-oriented strategies (e.g., outsourcing, explanation-seeking, and verification-seeking) which together constitute comparatively weak epistemic processes. Overall, 78.8% of student turns do not exhibit clear mastery-oriented aims and lack indicators of reliable epistemic strategies. This distribution points to generally low epistemic engagement and helps explain why apparent task success may have limited correspondence with genuine learning. In contrast, the sixth and seventh most frequent profiles reveal a distinct pattern in which mastery-oriented epistemic aims are strongly associated with epistemic judgment and, to a lesser extent, explanation-seeking strategies. Although these profiles account for only 11.1% of observed interactions, they suggest that when learners explicitly adopt mastery-oriented aims, they are more likely to engage in more reliable epistemic processes, particularly those involving strategies of justification during interaction with LLM assistants. These results underscore the necessity of EAIL, as they reveal how students may perform pseudo-success when completing tasks through superficial LLM interactions without engaging in the deep epistemic processing required for genuine learning. Identifying specific epistemic indicators within the EAIL framework, it is possible to better distinguish between this hallucination of learning and genuine learning, highlighting a critical need for CS curricula that prioritize high-level epistemic engagement over mere task completion.

## 4. Conclusion and Limitations

This study makes three primary contributions to research on AI literacy and AI-supported learning. Theoretically, it introduces EAIL as a process-oriented conceptual framework and demonstrates its applicability in analyzing real student–AI dialogue data. By operationalizing epistemic aims and processes at the level of student–AI interaction, this study extends AI literacy research beyond self-report and trait-like measures and introduces a principled process-level lens for examining epistemic engagement during authentic human–AI inquiry. Methodologically, the study implements a human-in-the-loop LLM-facilitated automatic labeling pipeline. Results show that integrating regex-based linguistic rules into few-shot prompting improves labeling accuracy compared to unconstrained few-shot prompting alone, highlighting the value of hybrid symbolic–LLM approaches. Empirically, the findings reveal that students are generally unlikely to engage in high levels of epistemic AI literacy during AI-assisted programming tasks. The most frequent epistemic profiles indicate that learners often fail to spontaneously adopt mastery-oriented epistemic aims and instead rely on strategies such as outsourcing, explanation-seeking, and verification-seeking, which together constitute comparatively weak epistemic processes. Although explanation-seeking and verification-seeking are important components of learning, reliance on these strategies alone does not necessarily reflect epistemically productive engagement, as they may occur without learner agency in evaluating, justifying, or regulating AI-generated information. In contrast, a smaller subset of interactions exhibits higher-quality epistemic engagement, characterized by mastery-oriented aims and the use of advanced strategies (e.g., epistemic justification). These patterns suggest that when learners explicitly orient toward long-term understanding and mastery, they are more likely to engage in more reliable epistmeic processes for inquiry with AI and genuine learning.

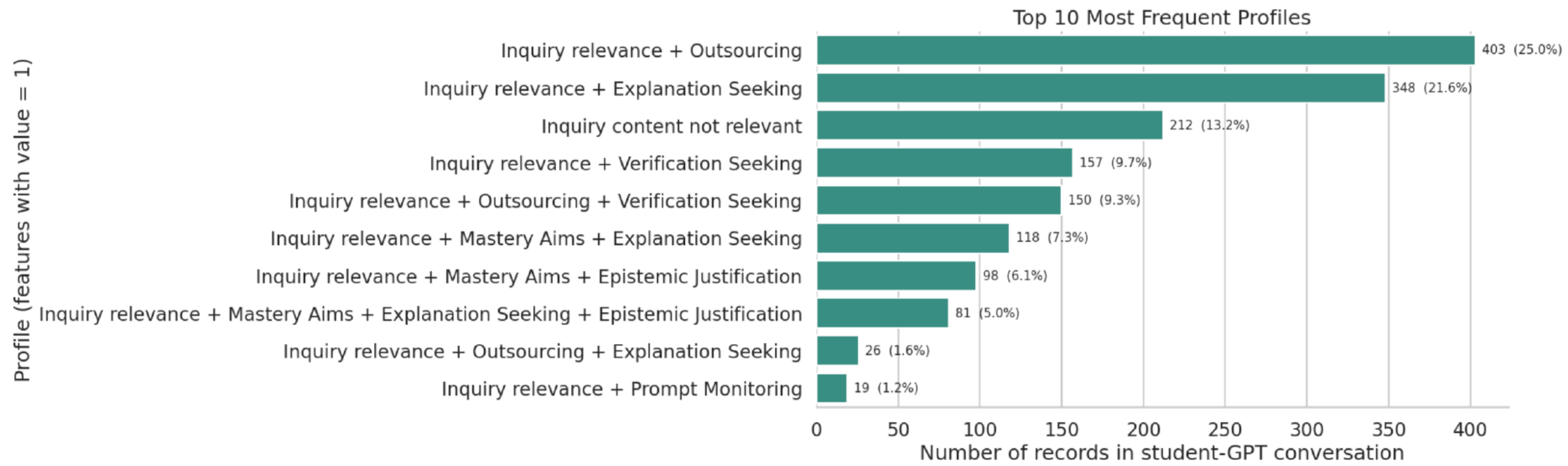

**Figure 4:** Ranking of the top ten learner profiles based on EAIL performance. Building on the data in Figure 3 (b), this chart illustrates the relative frequency of each profile pattern in descending order.

Several limitations and future directions should be noted. First, the size and scope of the analyzed dataset could be expanded as resources permit, enabling stronger generalization and re-validation of observed epistemic patterns. Second, future work could compare additional automatic labeling approaches and evaluate multiple LLMs within the same few-shot prompting and pipeline design to examine robustness across models. These findings carry important implications for the design of AI-supported learning environments. In unregulated settings, surface-level task success and pseudo-success may become the dominant mode of interaction, as robust epistemic engagement does not reliably emerge without intentional instructional and design support. This raises critical questions about how learners perceive the epistemic authority of LLMs and how responsibility for judgment, trust, and evaluation is distributed between humans and AI systems.

## A. Data and Code Availability

The code and materials for this study are available via

- GitHub repository.

## Declaration on Generative AI

During the preparation of this work, the author(s) used ChatGPT, Gemini, and Claude in order to: Grammar and spelling check, Paraphrase and reword, Improve writing style. The author(s) also used an AI-assisted coding tool within the Prism LaTeX environment to assist with code generation and debugging for the analysis pipeline described in Section 2. After using these tool(s)/service(s), the author(s) reviewed and edited the content as needed and take(s) full responsibility for the publication's content.